\def\year{2021}\relax
\documentclass[letterpaper]{article} 
\usepackage{aaai21}  
\usepackage{times}  
\usepackage{helvet} 
\usepackage{courier}  
\usepackage[hyphens]{url}  
\usepackage{graphicx} 
\usepackage{natbib}  
\usepackage{caption} 
\newcommand{\tablestyle}[2]{\setlength{\tabcolsep}{#1}\renewcommand{\arraystretch}{#2}\centering\footnotesize}

\graphicspath{{./images/}}
\DeclareGraphicsExtensions{.pdf}

\def \etal {{\emph{et al}.\thinspace}}

\def \eg {{\emph{e.g}.\thinspace}}
\def \ie {{\emph{i.e}.\thinspace}}

\def \shortcite {\cite}

\def \mp {\mathbf{p}}

\def \mv {\mathbf{v}}
\def \ms {\mathbf{s}}

\usepackage{amsmath}
\usepackage{amsfonts}
\usepackage{amssymb}
\usepackage{mathtools}
\usepackage{siunitx}  
\usepackage{overpic}
\usepackage{booktabs}
\usepackage{cleveref} 
\usepackage{comment}

\usepackage[linesnumbered,ruled,vlined]{algorithm2e}
\SetKwInput{KwOutput}{Output}              
\SetKwInput{KwInput}{Input}                

\renewcommand{\paragraph}[1]{\vspace{1mm}\noindent\textbf{#1.}}

\usepackage{wrapfig}
\usepackage{multirow}
\usepackage{makecell}
\usepackage{enumitem}
\usepackage{microtype}

\usepackage[switch]{lineno}  %

\title{Unsupervised 3D Learning for Shape Analysis via \\ Multiresolution Instance Discrimination}
\author {
    Peng-Shuai Wang\textsuperscript{\rm 1},
    Yu-Qi Yang\textsuperscript{\rm 2,1},
    Qian-Fang Zou\textsuperscript{\rm 3,1},
    Zhirong Wu\textsuperscript{\rm 1},
    Yang Liu\textsuperscript{\rm 1},
    Xin Tong\textsuperscript{\rm 1} \\
}
\affiliations {
    \textsuperscript{\rm 1}Microsoft Research Asia \\
    \textsuperscript{\rm 2}Tsinghua University \\
    \textsuperscript{\rm 3}University of Science and Technology of China\\
}

\begin{document}

\maketitle


\begin{abstract}
We propose an unsupervised method for learning a
generic and efficient shape encoding network for different shape analysis
tasks. Our key idea is to jointly encode and learn shape and
point features from unlabeled 3D point clouds. For this purpose, we adapt
HRNet to octree-based convolutional neural networks for jointly encoding
shape and point features with fused multiresolution subnetworks and design a
simple-yet-efficient \emph{Multiresolution Instance Discrimination} (MID) loss
for jointly learning the shape and point features. Our network takes a 3D
point cloud as input and output both shape and point features. After training,
Our network is concatenated with simple task-specific back-ends and
fine-tuned for different shape analysis tasks. 
We evaluate the efficacy and generality of our method
with a set of shape analysis tasks, including shape
classification, semantic shape segmentation, as well as shape registration
tasks. With simple back-ends, our network demonstrates the best performance
among all unsupervised methods and achieves competitive performance to
supervised methods. 
For fine-grained shape segmentation on the PartNet dataset, our method even
surpasses existing supervised methods by a large margin. 
\end{abstract}

\section{Introduction}
\label{sec:intro}

3D shape analysis plays an important role in many graphics and vision applications. A key step in all shape analysis tasks is to extract representative features (or called descriptors) in different levels from 3D shapes. In particular, distinguishable shape instance features are preferred for shape classification, while per-point features are essential to fine-level analysis tasks, like semantic shape segmentation and registration.

Early methods compute handcrafted features of 3D shapes. Although these manually-designed features can preserve some good properties such as transformation invariant, they are difficult to be tailored to specific shape analysis applications. State-of-the-art methods integrate the feature extraction with specific shape analysis task and learn an end-to-end deep neural network with the supervision of labeled shape analysis results. The success of these supervised learning methods is built upon large-scale labeled datasets, and the networks optimized for one task are difficult to adapt to others.

Unsupervised pre-training methods first learn a feature extraction backbone network from an unlabeled dataset via carefully designed unsupervised pretext task losses. After that, the pre-trained backbone network is concatenated with task-specific back-end networks and refined for different downstream tasks via transfer learning. In computer vision and natural language processing tasks, unsupervised pre-training have demonstrated their advantages for reducing the workload of data labeling and network design \cite{Wu2016,Yang2018a,Deng2018,Zhao2019,Hassani2019}. However, these networks and training schemes cannot be easily adapted for 3D shape analysis due to irregular representation of 3D point clouds and multi-level shape features required by different shape analysis tasks. A set of unsupervised 3D learning methods \cite{Wu2016,Yang2018a,Deng2018,Zhao2019} have been proposed for extracting shape features from 3D point clouds, none of them offers a generic backbone network for different shape analysis tasks with competitive performance to the supervised methods. 

In this paper, we present an unsupervised pre-training method for learning a generic 3D shape encoding network for 3D shape analysis. Our key observation is that a 3D shape is composed of its local parts and thus the feature for shape and points are coherent and should be encoded and trained jointly. Based on this observation, our shape encoding backbone network adapts HRNet \cite{Wang2019b} to an octree-based convolutional network~\cite{Wang2017} for extracting and fusing features from both points and shapes via parallel multiresolution subnetworks and connections across subnetworks. It takes 3D point cloud as input and outputs an instance-wise feature of the whole 3D shape as well as point-wise features. Inspired by the instance discrimination designed for 2D image classification \cite{Wu2018a}, we design a simple-yet-efficient Multi-resolution Instance Discrimination (MID) losses for supervision of extracted shape and point features, in which a shape instance discrimination loss classifies augmented copies of each shape instance of a 3D dataset in one class, while a point instance discrimination loss classifies the same points on the augmented copies of a shape instance in a class.


We trained our backbone shape encoding network (denoted as MID-Net) with ShapeNetCore55 \cite{Chang2015} and evaluated its performance with simple back-ends in various shape analysis tasks, including shape classification, two shape segmentation tasks, and 3D shape registration. Our experiments demonstrate that in all these tasks, our pre-trained backbone offers better performance than the same network trained with the labeled data of downstream tasks, especially as the amount of labeled data in the downstream tasks becomes small. Among all unsupervised 3D learning methods, our method achieves the best performance in all shape analysis tasks. Moreover, it achieves competitive performance to the state-of-the-art supervised methods in all tasks. In fine-grained PartNet segmentation, our method surpasses state-of-the-art supervised methods by a large margin. 

\section{Related Work}
\label{sec:related}

\paragraph{Supervised 3D feature learning}
Discriminative shape features can be learned with the supervision of the labeled
data 
in a task-specific manner. Supervised deep learning approaches often achieve the
best performance on the datasets of shape classification and segmentation
\cite{Yi2017,Mo2019,Yu2019}. Existing 3D deep learning methods can
be classified according to shape representations: multi-view based CNNs
\cite{Maturana2015,Su2015,Choy2016,Kalogerakis2017}, volumetric and
sparse-voxel-based CNNs
\cite{Wu2015,Graham2015,Wang2017,Riegler2017,Graham2018}, point-based networks
\cite{Qi2017,Qi2017a,Li2018,Li2018a}, manifold-based CNNs
\cite{Boscaini2015,Boscaini2016,Hanocka2019} and graph-based approaches
\cite{Yi2017b,Wang2019c}. Despite the good performance, the task-specific
learned features are difficult to adapt to other tasks and preparing labeled 3D
data is tedious and costly.

\paragraph{Unsupervised 3D feature learning}
unsupervised learning methods aim to learn generic 3D shape representations from
unlabeled data via carefully designed pretext tasks.

3D-GANs~\cite{Wu2016} train a generative adversarial network (GAN) on volumetric
data, and its discriminator is used for extracting shape-level features. L-GANs
learn deep shape representations by combining an autoencoder network and a GAN
~\cite{Achlioptas2018}. FoldingNets~\cite{Yang2018a} and
AtlasNets~\cite{Groueix2018} optimize an autoencoder by reconstructing shapes
from deformed point clouds. Zhang and Zhu \shortcite{Zhang2019a} cascade two
pretext tasks --- part contrasting and object clustering to learn the shape
feature space. The unsupervised shape-level features are mainly used for shape
classification and retrieval. 

Deng \etal \shortcite{Deng2018} and Shu \etal \shortcite{Shu2016} first extract
hand-crafted point-pair features from geometric patches, then uses an
autoencoder to compress the features. SO-Nets~\cite{Li2018a} extract
hierarchical features from individual points and nodes of a self-organizing map.
PointCapsNets~\cite{Zhao2019} extend capsule networks~\cite{Sabour2017} to 3D
point clouds and are trained by the shape reconstruction loss. Hassani and
Haley~\shortcite{Hassani2019} propose to train a multiscale graph-based encoder
with multiple pretext tasks. Li \etal~\shortcite{Li2020} propose an unsupervised
clustering task to learn point features for detecting distinctive shape regions.
The work~\cite{Sauder2019} proposes a novel pretext task that reconstructs the
voxel indices of randomly arranged points. 
Concurrently, Xie \etal~\shortcite{Xie2020} use shape registration as the
pretext task and adopts the point contrastive loss to learning point features;
Yang \etal~\shortcite{Yang2020} design a pretext task of finding correspondences
between shapes belonging the same category based on the cycle-consistency.

Although the above works can generate both point-level and shape-level features,
their pretext tasks do not impose explicit self-supervision on both levels. We
fill this gap by discriminating different-level features simultaneously and
achieve significant improvements.

\paragraph{Unsupervised pre-training}
Unsupervised pre-training is actively studied due to its success in various
fields. In the natural language processing field, BERT models~\cite{Devlin2018}
use masked context prediction and next sentence prediction as the pretext tasks,
and  GPT models~\cite{Radford2018} prefer the language modeling task. In
computer vision field, various pretext tasks or dedicated loss functions like
colorization~\cite{Zhang2016}, context prediction~\cite{Doersch2015}, motion
segmentation~\cite{Pathak2017}, iterative feature clustering~\cite{Caron2018},
image instance discrimination~\cite{Wu2018a} and contrastive
losses~\cite{Hadsell2006,He2020}, have also shown their strengths in learning
effective image features.

Inspired by instance discrimination~\cite{Wu2018a} which operates on a single
image level, we propose to discriminate multiresolution instances of shapes and
develop effective training schemes to reduce the computational and memory cost
caused by the large number of instances which cannot be handled by Wu \etal's
approach~\shortcite{Wu2018a}.

\section{Method}
\label{sec:method}

\begin{figure*}[t]
\centering
\begin{overpic}[width=0.98\linewidth]{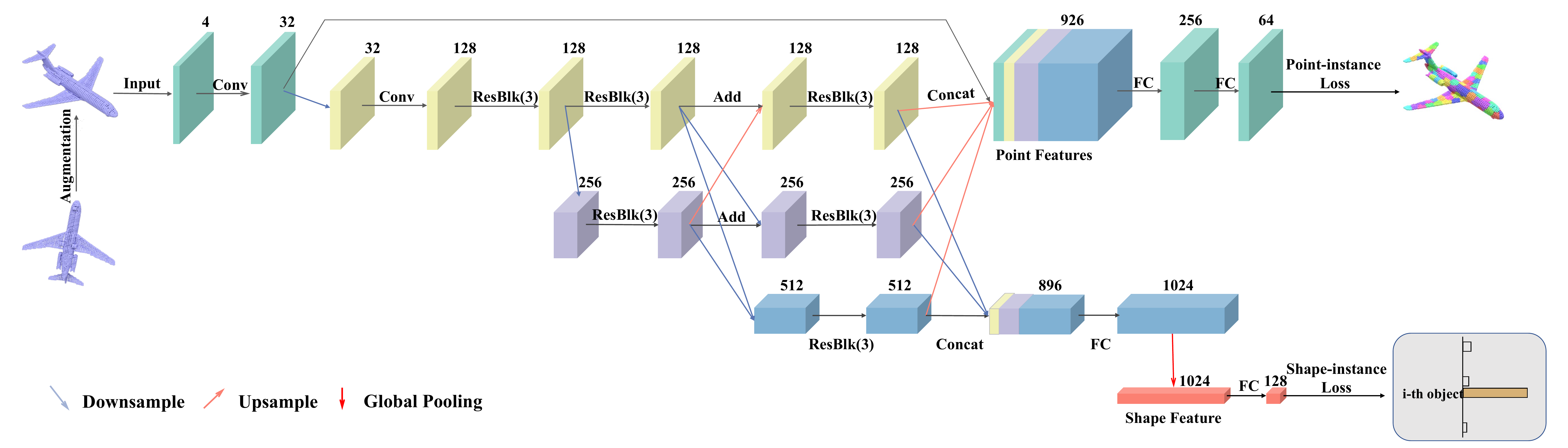}
\end{overpic}
\vspace{-10pt}
\caption{Overview of our multiresolution-instance-discrimination (MID) unsupervised pre-training pipeline.   An augmented input point cloud via transformations is fed into the deep neural network which maintains and fuses multi-scale resolution feature maps. The shape-level features and point-wise features are extracted from the network, and they are encouraged to be discriminative and transformation-invariant under the supervision of the MID loss on both shape instance and point levels. }
\label{fig:overview}
\end{figure*}

Given a large unlabeled 3D shape collection that consists of $N$ 3D models,
we assume each 3D shape $X_i$ is represented with a point cloud with $M_i$ points, and denote the $j$-th point of $X_i$ by $\mp_{i,j}$. Here the term ``unlabeled'' means that the data has no shape category information or other human-annotated labels. Our goal is to train a shape encoder network $f_{\theta}$ that takes the point cloud of $X_i$ as input and generate the representative and  shape-level feature $\ms_i$ and point-wise features $\{\mv_{i, j}, j=1,\ldots, M_i\}$ for $X_i$:
\begin{equation}
f_{\theta}: X_i \longmapsto [\ms_i, \mv_{i, 1}, \cdots, \mv_{i, M_i}],
\end{equation}
where $\theta$ denotes the set of network parameters.

To train the above feature space in an unsupervised manner, we first augment shapes via various transformations and create multiresolution class labels for later training (see \Cref{subsec:data}), then feed them into a deep neural network which maintains and fuses multi-scale resolution feature maps efficiently (see \Cref{subsec:network}). We use the multi-resolution instance discrimination pretext task (see \Cref{subsec:loss}) to self-supervise the feature learning progress (see \Cref{subsec:training}). Figure~\ref{fig:overview} shows the overview of our method.

\subsection{Input Data Processing}
\label{subsec:data}
\paragraph{Preprocessing and data augmentation}
We pre-process the input point cloud to assign point normals via principal component analysis if the accurate normal information is not available. For data augmentation, we first normalize each point cloud inside
a unit sphere, then generate shape instances with a transformation composed by
random rotations, random translations within [-0.25, 0.25], and random scaling
along each coordinate axis with the ratio within [0.75,1.25]. For the dataset
in which the up-right directions of models are aligned, our random rotation is restricted on the rotation along the upright axis.   

\paragraph{Multiresolution instance class creation}
We label the transformed instances of the same shape with its index in the input dataset. Also, each point on the generated instance is labeled by the same index of the corresponding point in the input shape in the dataset. So in total,
We create $N$ shape-instance classes and $M_i$ point-instance classes for each shape-instance class. These multiresolution class labels serve as self-supervision signals for our network training. 
As the total number of point-level classes of a large-scale shape dataset could be huge: $\sum_{i=1}^N M_i$,  it will lead to huge memory consumption in network training. To overcome this  issue, we introduce the concept of patch-instance class. On each shape $X_i$ in the dataset, we over-segment it into $K_i$ patches ($K_i\ll M_i$), and on each shape-instance class, we can create $K_i$ patch-instance classes, similar to the construction of point-instance classes. In this way, the point-instance classes can be approximated by the patch-instance classes, which have a less total number. For simplicity, we use the K-Means algorithm to compute over-segmented patches and set $K_i=100$.

\subsection{Network Design}
\label{subsec:network}

We adopt HRNet ~\cite{Wang2019b} to extract multiresolution instance features. Different from the conventional U-Net~\cite{Ronneberger2015} architecture that cascades sequential convolutional layers from high to low resolutions and then recovers high resolution features from low-resolution ones in a reversed order, HRNet maintains parallel multiresolution subnetworks and simultaneously outputs multi-resolution features. The features extracted by subnetworks in different resolutions are fused in different intermediate stages of HRNet. For 3D shape encoding, HRNet can simultaneously output low-resolution shape-level features and high-resolution point-wise features with one network.  This property well matches our observation about multiresolution instance features as introduced in \Cref{sec:intro}.
Also, we can easily apply loss functions for outputs in different resolutions and each loss function will contribute to the training of all subnetworks.
We built HRNet upon an octree-based CNN framework~\cite{Wang2017} due to its efficiency in both computational cost and memory consumption and its natural multiresolution representation for building multiresolution subnetworks. A 3D point cloud is first converted to an octree representation, by default, in $64^3$ resolution.

The details of the HRNet structure used in our method are illustrated in \Cref{fig:overview}. The numbers of feature channels are listed on the top of each feature map. The convolution kernel size is $3\times 3 \times 3$. ResBlk(3) represents three cascaded ResNet blocks with a bottleneck structure~\cite{He2016}. The Downsample operation is implemented by max-pooling, and the Upsample operation  is the tri-linear up-sampling.

\subsection{Multi-Resolution Instance Discrimination}
\label{subsec:loss}
Our network training is supervised by two loss functions:  shape-instance discrimination loss and point-instance discrimination loss. Both loss functions are based on input 3D models and their multiresolution instance class labels created in \Cref{subsec:data}.

\paragraph{Shape-instance discrimination loss}
The shape-level feature is supervised by a linear classifier that classifies shape instances into $N$ classes, where each 3D shape $X_i$ and its augmented instances are classified into the $i$-th class. The cross-entropy loss for the $i$-th class is defined as:
\begin{equation}
L_s(\ms_i) = - \log
\dfrac{\exp((\ms_i \cdot \tilde{\ms}_i) / \tau_s)}
  {\sum_{k=1}^{N} \exp((\ms_i \cdot \tilde{\ms}_k) / \tau_s)},
\label{eq:shape}
\end{equation}
where $\tilde{\ms}_i$ is the weight vector in the linear classifier for the $i$-th class. We follow the approach of \cite{Wang2018c,Wu2018a} to normalize $\ms_i$ and $\tilde{\ms}_i$ to be unit-length and measure their difference by the cosine distance. $\tau_s$ is a parameter controlling the concentration level of the extracted features and is set to $0.1$ empirically.

To obtain a good classification, an optimal $\tilde{\ms}_i$ expects to be the average center of features of all shape instances in this class \cite{Liu2018b}. By optimizing the loss function \Cref{eq:shape}, the features of shape instances under different transformations converge to $\tilde{\ms}_i$ and thus they are invariant to the imposed transformations. Meanwhile, the loss function of \Cref{eq:shape} tends to maximize the distance between $\ms_i$ and other $\tilde{\ms}_k (k\neq i)$, which makes the shape feature $s_i$ of shape $X_i$ is discriminative against the features of other 3D models.

\paragraph{Point-instance discrimination loss}
To optimize point-wise features, we can also use a linear classifier to classify the points of a shape and its augmented copies into $M_j$ classes, with the created supervision signals (\Cref{subsec:data}). For points in the $j$-th point class, the cross-entropy loss is defined as:
\begin{equation}
L_{p}(\mv_{i, j}) = -\log
\dfrac{\exp((\mv_{i, j} \cdot \tilde{\mv}_{i, j}) / \tau_p)}
  {\sum_{k=1}^{M_i} \exp((\mv_{i, j} \cdot \tilde{\mv}_{i, k}) / \tau_p)},
\label{eq:point}
\end{equation}
where $\mv_{i, j}$ is the point-wise feature of $j^{th}$ point of an augmented shape instance of $X_i$, $\tilde{\mv}_{i,j}$ is the weight vector in the linear classifier for the $j$-th point class. All the point feature vectors are also unit-length and the control parameter $\tau_p$ is set to 0.1.

As discussed in \Cref{subsec:data}, treating each point of a 3D shape as an individual class is impractical as it leads to a large number of classes for each object and results in huge memory consumption and computational cost for storing and updating the weights of linear classifiers. Therefore, we propose to approximate point-instance discrimination by patch-instance discrimination, and revise the loss function as:
\begin{equation}
L_{p}(\mv_{i, j}) = -\log
\dfrac{\exp((\mv_{i, j} \cdot \tilde{\mv}_{i, c(i,j)})/ \tau_p)}
  {\sum_{c=1}^{K} \exp((\mv_{i, j} \cdot \tilde{\mv}_{i, c}) / \tau_p)},
\label{eq:patch}
\end{equation}
where $c(i, j)$ is the index of  the patch containing  the $j^{th}$ point on shape $X_i$.

\paragraph{MID Loss}
By combining the above two multiresolution instance discrimination loss functions,  we define the MID loss for an instance of shape $X_i$ as:
\begin{equation}
L(X_i) = L_{s}(\ms_i) +
 \sum\nolimits_{j=1}^{M_i} L_{p}(\mv_{i, j}) / M_i.
\label{eq:loss}
\end{equation}
Note that although our method applies different linear classifiers for points of different 3D shapes, the shape encoder is shared by all 3D shapes.

\subsection{Network Training}
\label{subsec:training}
Given the MID loss function defined above, we optimize the network parameters of $f_\theta$ with an input 3D shape collection by a stochastic gradient descent (SGD) algorithm. In each mini-batch, we randomly pick a set of 3D shapes from the dataset and generate an augmented shape for each original shape in runtime.

\begin{algorithm}[b]
\small
\DontPrintSemicolon
\KwInput{A set of shapes $\{X_i\}_{i=1}^N$;}
\KwOutput{Network $f_\theta$, $\{\tilde{s}_i\}_{i=1}^N$ and $\{ \{\tilde{v}_{i, c}\}_{c=1}^K \}_{i=1}^N$; }
Assign multiresolution instance labels to $\{X_i\}_{i=1}^N$ according to \Cref{subsec:data} \\
\For{ l in [0, max\_iteration]} {
    Randomly sample a batch of $B$ samples $\{X_{b_i}\}_{i=1}^B$; \\
    Compute loss  $L = \frac{1}{B} \sum_{i=1}^{B} L(X_{b_i})$ (\Cref{eq:loss});\\
    Compute gradient $\nabla L(\theta)$ and update $\theta$ with SGD; \\
    Update $\{\tilde{\ms}_{b_i}\}_{i=1}^B$ and $\{ \{\tilde{v}_{{b_i}, c}\}_{c=1}^K \}_{i=1}^B$ with \Cref{eq:update_vi} \& \Cref{eq:update_pi}.
  }
\caption{Network training procedure}
\label{algorithm}
\end{algorithm}

A na\"{i}ve approach to train the network is to update the weights of the shape encoder network and the linear classifier together according to the back-propagated gradient of the MID loss function. However, we found that updating the linear classifiers in each mini-batch often makes the classifier training unstable and hinders the optimization of the shape encoder. 
 Inspired by the temporal ensembling scheme in ~\cite{Laine2017}, we update the weights of the linear shape-instance classifier slowly by
\begin{equation}
\tilde{\ms}_i  = (1 - \lambda_s) \cdot \tilde{\ms}_i + \lambda_s \cdot \ms_i,
\label{eq:update_vi}
\end{equation}
where $\ms_i$ is the shape-level feature of the current shape instance for the $i$-th shape-instance class, $\lambda_s$ is a momentum parameter and is set to 0.5 in our implementation. For point-instance classifier, we use a similar update rule:
\begin{equation}
\tilde{\mv}_{i,c} = (1 - \lambda_p) \cdot \tilde{\mv}_{i,c} + \lambda_p \cdot \mv_{i,c},
\label{eq:update_pi}
\end{equation}
where $\mv_{i,c}$ is the average of the point-wise features of all points that belong to the patch class $c$, $\lambda_p$ is a momentum parameter and is set to 0.5 too.
A detailed training procedure is summarized in Algorithm~\ref{algorithm}.

\section{MID-Nets for Shape Analysis}
\label{sec:results}
In this section, we first present our back-end network design and its training
scheme and then discuss the performance of our method in each downstream task.

\subsection{Backend Design and Training}
For each downstream task, we concatenate MID-Net with simple back-end layers. We
use a one-layer fully-connected (FC) network for shape classification and
two-layer FC for shape segmentation. We denote the MID-Net with the FC back-end
as \textbf{MID-FC} and optimize the concatenated networks with two training
schemes:
\begin{enumerate}[leftmargin=*]\setlength\itemsep{0mm}
    \item[-] \textbf{MID-FC(Fix)}: we fix the pre-trained MID-Net and only train
    the back-end with the labeled training data in each shape analysis task.
    \item[-] \textbf{MID-FC(Finetune)}: we fine-tune both MID-Net and the FC
    back-end with the labeled training data in each shape analysis task. The
    MID-Net is initialized with the pre-trained weights, the FC back-end is
    randomly initialized.
\end{enumerate}
To evaluate the impact of pre-training, we also randomly initialize both MID-Net
and FC layers and train the network from scratch, denoted as
\textbf{MID-FC(NoPre)}.

We trained our MID-Net on ShapeNet dataset~\cite{Chang2015} that consists of
57,449 3D shapes. The overall network parameter size is 1.5M, which is
comparable to PointNet++~\cite{Qi2017} (1.4 M), DGCNN~\cite{Wang2019c} (1.5 M),
and much smaller than PointCNN~\cite{Li2018} (8.2 M). The detailed training
hyper-parameters are provided in the supplemental material.

\subsection{Shape Classification}

\paragraph{Dataset}
We use the ModelNet40~\cite{Wu2015}, which contains 13,834 3D models across 40
categories: 9,843 models are used for training and 3,991 models for testing. We
use the classification accuracy as the evaluation metric.

\paragraph{Results and comparisons}
\Cref{tab:finetune} lists the testing accuracy on the ModelNet40. For a
fair comparison,  we do not use the voting or orientation pooling
strategy~\cite{Qi2017} to improve the results. As shown in the
first column, our pre-trained MID-Net provides a good initialization to
MID-FC(Finetune) and thus results in better accuracy (93.1\%) than MID-FC(NoPre)
(92.9\%). Compared to the state-of-the-art supervised methods including
PointNet++~\cite{Qi2017}, PointCNN~\cite{Li2018}, DGCNN~\cite{Wang2019c},
KPConv~\cite{Thomas2019}, and RS-CNN~\cite{Liu2019}, our  MID-FC(Finetune) also
exhibits superior accuracy.

We also compare our method with other unsupervised methods  including
L-GAN~\cite{Achlioptas2018}, FoldingNet~\cite{Yang2018a},
Multi-Task~\cite{Hassani2019}, PointCapsNet~\cite{Zhao2019},  and
ReconSpace~\cite{Sauder2019}. Here all methods are pre-trained
with the ShapeNet dataset and use one FC layer, \ie, a linear classifier. Our
MID-FC(Fix) achieves the second-best performance (90.3\%) among all listed
unsupervised approaches, and is slightly worse than
ReconSpace.

In \Cref{tab:cls:finetune2}, we evaluate the accuracy of MID-FC learned from
different ratios of labeled data, where our method attains good performance and
is better than  FoldingNet~\cite{Yang2018a}. MID-FC(Finetune) further improves
the classification accuracy and verifies the effectiveness of our pretraining.

\begin{table}[t]
\centering
\tablestyle{2pt}{1.0}
\scalebox{0.9}{%
    \begin{tabular}{lc|lc|lc}
    \toprule
    Our Method      & Accu.         & Supervised   & Accu.    & Unsupervised      & Accu.     \\
    \midrule
    MID-FC(NoPre)    & 92.9            & PointNet++   & 90.7        &  L-GAN            & 84.5         \\
    MID-FC(Fix)      & 90.3            & PointCNN     & 92.2        &  FoldingNet       & 88.4        \\
    MID-FC(Finetune) & \textbf{93.0}   & DGCNN        & 92.9        &  Multi-Task       & 89.1         \\
                     &                 & KPConv       & 92.9        &  PointCapsNets    & 89.3         \\
                     &                 & RS-CNN       & 92.9        &  ReconSpace        & 90.6         \\
    \bottomrule
    \end{tabular}
  } 
  \vspace{-5pt}
  \caption{ModelNet40 shape classification. We list the performance of the state-of-the-art supervised
  methods (2nd column) and unsupervised methods (3rd column). }
\label{tab:finetune}  
\end{table}

\begin{table}[t]
\tablestyle{4pt}{1.0}

\scalebox{0.9}{%
    \begin{tabular}{lccccccc}
    \toprule
    Data            & 1\%   & 2\%    & 5\%    & 10\%  & 20\% \\ 
    \midrule
    FoldingNet      & 56.4  & 66.9   & 75.6   & 81.2  & 83.6 \\ 
    MID-FC(Fix)     & \textbf{61.5}  & \textbf{73.1}   & \textbf{80.2}   & \textbf{84.2}  & \textbf{86.9} \\ 
    \midrule
    MID-FC(NoPre)   & 58.5  & 71.2   & 80.1   & 85.4  & 88.7 \\
    MID-FC(Finetune)& \textbf{67.3}  & \textbf{76.5}   & \textbf{83.6}   & \textbf{88.4}  & \textbf{90.2} \\
    \bottomrule
    \end{tabular}
  } 
  \caption{The comparison of MID-FC(Fix) and FoldingNet for shape classification with a linear classifier as the back-end. 
  }
\label{tab:cls:finetune2}
\end{table}

\begin{table}[t]
\centering
\tablestyle{4pt}{1.1}

\scalebox{0.95}{
\begin{tabular}{lc|lc}
\toprule
Methods             & mIoU            & Methods        & mIoU  \\
\midrule
MID-FC(NoPre)       & 58.4            & PointNet++     & 49.6  \\
MID-FC(Fix)         & 49.4            & SpiderCNN      & 48.7  \\
MID-FC(Finetune)    & \textbf{60.8}   & PointCNN       & 47.9  \\
\bottomrule
\end{tabular}
}
\vspace{-5pt}
\caption{The semantic segmentation mIoU on PartNet. }
\label{tab:partnet}
\end{table}

\begin{table}[t]
    \centering
\tablestyle{4pt}{1.1}

    \scalebox{0.95}{
    \begin{tabular}{lrrrrr}
    \toprule
    Model & 1\% & 2\% & 5\% & 10\% & 20\% \\
    \midrule
    MID-FC(NoPre) & 14.5  & 16.1 & 27.7 & 29.5 & 41.6 \\
    MID-FC(Fix)   & 21.7  & 25.2 & 31.3 & 34.4 & 38.9 \\
    MID-FC(Finetune) & \textbf{21.8} & \textbf{26.8} & \textbf{35.4} & \textbf{40.2} & \textbf{46.3} \\
    \bottomrule
    \end{tabular}
    }
    \vspace{-5pt}
    \caption{The semantic segmentation mIoU on PartNet with different ratios of labeled data. }
   \label{tab:partnet2} 
   \vspace{-10pt}
\end{table}
    
\subsection{Fine-Grained PartNet Segmentation}
\paragraph{Dataset}
We evaluate our method with fine-grained segmentation on the PartNet 
\cite{Mo2019}, which is a challenging dataset that consists of 24,506 3D shapes
in 17 categories with fine-grained (\ie, Level-3) labels. Each shape contains
10,000 points and the part numbers in each shape category vary from 3 to 50.  We
follow the data split setup in \cite{Mo2019}.  We use the mean IOU across all 17
categories as the evaluation metric.   
Note that the ShapeNet dataset used for
our pre-training does not contain Door, Fridge, and Storage categories of
PartNet.

\paragraph{Results and comparisons}
\Cref{tab:partnet} shows the performance of MID-FC and other state-of-the-art
supervised approaches, including PointNet++~\cite{Qi2017},
SpiderCNN~\cite{Xu2018}, and PointCNN~\cite{Li2018}.
For fair comparisons, we trained each model with the code provided by the
original authors on the same dataset. 
And during the testing phase, we did not use the voting strategy to
improve the result. Our MID-FC(Fix) outperforms most existing supervised
approaches and is only slightly worse than PointNet++ by 0.2, which is
significant considering the fact that MID-FC(Fix) only contains 2 trainable FC
layers and optimizes less than 0.3M parameters. With fine-tuning,
MID-FC(Finetune) achieves more  improvements and the accuracy increases
\emph{11.2} points at least compared to other supervised approaches, and 2 point
improvement over MID-FC(NoPre). The results clearly show the benefit of our
pre-training.  For the three categories that are only contained in PartNet while
not in ShapeNet, MID-FC(Finetune) and MID-FC(Fix) still achieve better
performance (see the supplemental material), which demonstrates the good
generality of our pre-trained features.

The segmentation performance of all our three networks learned from different
ratios of training data  is reported in \Cref{tab:partnet2}. MID-FC(Finetune)
and MID-FC(Fix) have better performance than the supervised MID-FC(NoPre), and
the accuracy of MID-FC(Fix) and MID-FC(Finetune) trained with 20\% labeled data
is comparable to other supervised methods trained with 100\% labeled data.

\begin{table}[t]
\centering
\tablestyle{1pt}{1.0}
\scalebox{0.88}{%
  \begin{tabular}{lcc||lcc}
    \toprule
    No Voting & C.mIOU  & I.mIOU  & Voting    & C.mIOU  & I.mIOU \\
    \midrule            
    PointNet++        & 81.9    & 85.1    & Submanifold       & 83.3    & 86.0 \\
    DGCNN             & 82.3    & 85.1    & RS-CNN            & 84.0    & 86.2 \\
    SynSpecCNN        & 82.0    & 84.7    & PointCNN          & 84.6    & 86.1 \\
    SPLATNet          & 83.7    & 85.4    & KPConv            & 85.1    & \textbf{86.4} \\
    MID-FC(NoPre)     & 84.1    & 85.2   & MID-FC(Finetune)        &  \textbf{85.2} & 85.8  \\
    MID-FC(Fix)       & 82.8    & 84.1   &         &   &    \\
    MID-FC(Finetune)  & \textbf{84.3} & \textbf{85.5} &   &  &   \\
    \bottomrule
  \end{tabular}
  }
  \vspace{-5pt}
\caption{The segmentation mean IoU across all categories (C.mIoU) and all instances (I.mIoU) on the ShapeNet Part. }
\label{tab:shapenet_seg}
\end{table}

\begin{table}[t]
    \centering
\tablestyle{6pt}{1.0}
    
    \scalebox{0.9}{
    \begin{tabular}{lrr|rr}
    \toprule
    \multirow{2}{*}{Model} & \multicolumn{2}{c|}{C.mIoU} & \multicolumn{2}{c}{I.mIoU} \\
    \cmidrule{2-5}                       
                      & 1\%   & 5\%  & 1\%   & 5\%   \\
    \midrule   
    PointCapsNets     & -     & -    & 67.0  & 70.0    \\
    Multi-Task        & -     & 73.0 & 68.2  & 80.7    \\
    MID-FC(NoPre)     & 46.3  & 70.9 & 65.1  & 80.3    \\
    MID-FC(Fix)       & 66.2  & 76.5 & 72.4  & 80.9    \\
    MID-FC(Finetune)  & \textbf{67.6} & \textbf{77.8} & \textbf{76.2} & \textbf{82.1} \\
    \bottomrule
    \end{tabular}
    }
   \vspace{-5pt}

    \caption{The segmentation mIoU on ShapeNet Part with different ratios of labeled data. }
   \label{tab:shapenet_seg2} 
   \vspace{-10pt}
\end{table}

\subsection{ShapeNet Part Segmentation}

\paragraph{Dataset}
We use the ShapeNet Part dataset \cite{Yi2017} which contains 16,881 3D point
clouds, collected from 16 categories of ShapeNet. Each point cloud has 2 to 6
part labels.  The number of points of a shape varies from 1000 to 3000. We
follow \cite{Yi2017} to split the data.   We use the mean IoU across all
categories (C.mIoU) and the mean IoU across all instances (I.mIoU) as the
evaluation metrics.

\paragraph{Results and Comparisons}
\Cref{tab:shapenet_seg} shows the IoU of our methods and other state-of-the-art
methods, including supervised methods such as PointNet++~\cite{Qi2017},
SynSpecCNN~\cite{Yi2017b} and DGCNN~\cite{Wang2019c} and SPLATNet~\cite{Su2018},
Submanifold~\cite{Graham2018}, KPConv~\cite{Thomas2019}, PointCNN~\cite{Li2018},
RS-CNN~\cite{Liu2019}. On the left panel of \Cref{tab:shapenet_seg}, the methods
did not use the voting strategy during the testing stage, and on the right
panel, the methods used voting strategy during the testing stage. Compared with
these supervised methods, MID-FC(Finetune) have a better performance. In
\Cref{tab:shapenet_seg2}, we also compare our method with fine-tuned
unsupervised methods like Multi-Task~\cite{Hassani2019} and
PointCapsNet~\cite{Zhao2019} with 1\% and 5\% training data. Our MID-FC(Fix) and
MID-FC(Finetune) also achieves better performance. And the improvements of our
MID-FC(Finetune) over MID-FC(NoPre), especially when the training data
is limited, clearly demonstrates the advantage of our pre-trained MID-Net.

\subsection{Shape Registration}
The process of point cloud registration is to align a point cloud with its
transformed version. To handle arbitrary rotations of shapes,  we trained a new
MID-Net with shapes augmented with arbitrary rotations. We regard the point-wise
features extracted from our pre-trained MID backbone network as the point
descriptors and use them to find closet point pairs from two input point clouds
for computing the initial rigid transformation for the standard ICP algorithm.

\paragraph{Dataset}
We conduct a comparison on the test set of ModelNet40~\cite{Wu2015}, which
contains 2,468 man-made models across 40 categories. We normalize each shape
inside a unit sphere and apply a random rigid transformation to the shape. In
particular, the rotation angle along each coordinate axis is randomly sampled
from $[\ang{0}, \ang{360}]$, while the translation along each coordinate axis is
randomly sampled from $[-0.25, 0.25]$. We evaluate the registration results by
computing the Hausdorff distance between two registered shapes by different
algorithms with the same ICP refinement. 

\begin{figure}[t]
    \centering
    \begin{overpic}[width=0.85\linewidth]{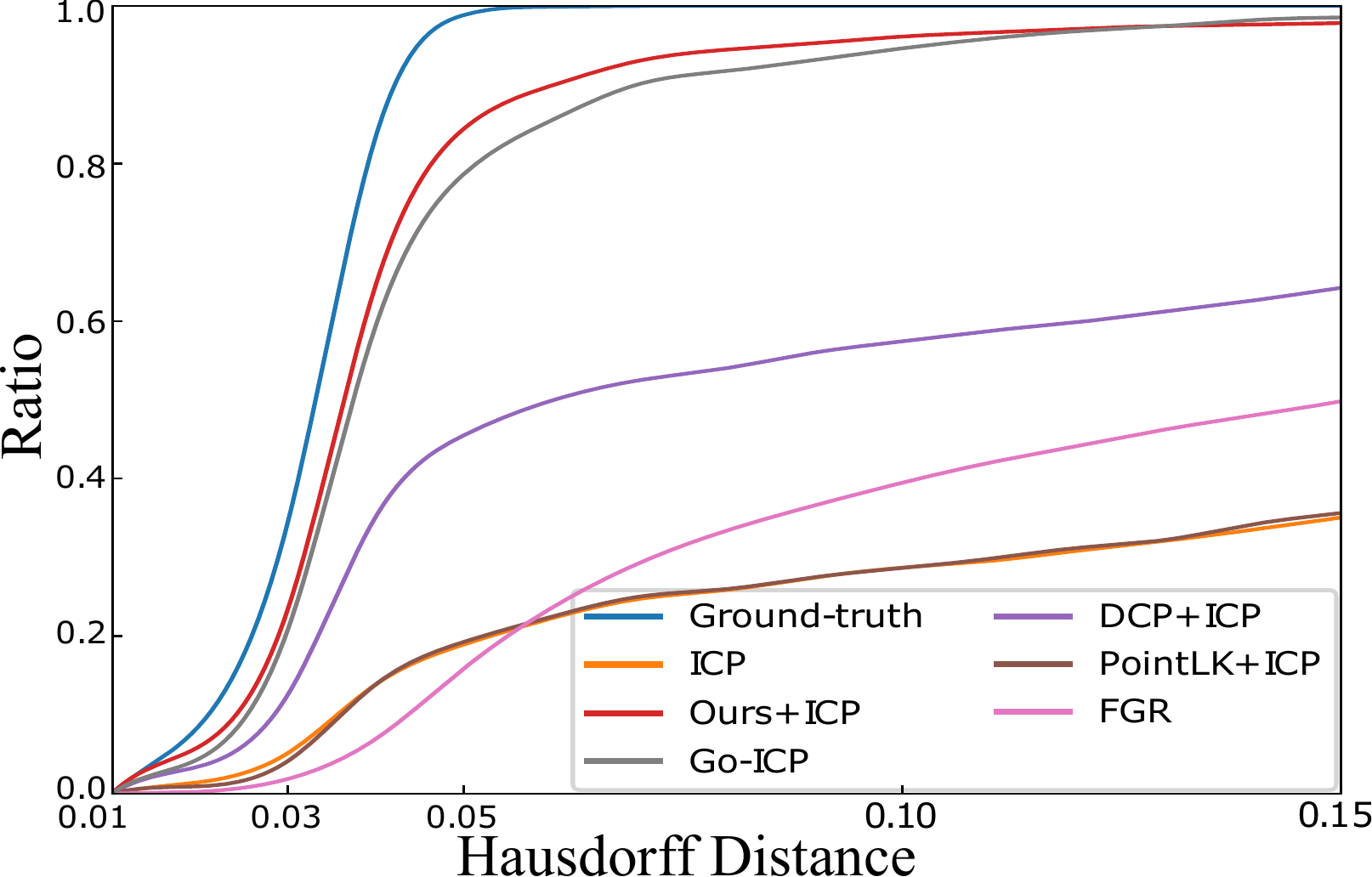}
    \end{overpic}
    \vspace{-5pt}
    \caption{Point cloud registration results. The curve represents the ratio of
    registration results under a specific Hausdorff distance. Our method
    achieves the best performance.} \label{fig:reg}
\end{figure}

\paragraph{Results and comparisons}
We compare our method with four state-of-the-art methods, including
Go-ICP~\cite{Yang2015}, FGR~\cite{Zhou2016}, PointLK~\cite{Aoki2019}, and
DCP~\cite{Wang2019a}.  DCP and PointLK were originally trained  with rotation
angles sampled within $[\ang{0}, \ang{45}]$. We re-trained their models with the
code provided by the authors. However, we found that the trained networks
perform poorly in dealing with large rotations. For all the methods except FGR
which performs nonlinear optimization, we run the ICP algorithm to refine the
initial registration. \Cref{fig:reg} shows the accuracy curve within a given
Hausdorff distance bound. The ground-truth curve and the curve by applying the
standard ICP algorithm serve as the upper and lower bound of all the algorithms.
Our method outperforms the other four methods. 

\section{Method Validation}
In this section, we first evaluate the contributions of our octree-based HRNet
and the MID loss, as well as the generality of our MID scheme for other network
architectures. After that, we conduct a set of ablation studies to validate
design decisions and parameter setting of our method.

\subsection{Contributions of Our Network and MID}
\paragraph{Contribution of our network}
We compare the performance of MID-FC(NoPre) with the existing supervised methods
that can achieve the best performance in each downstream task. 
Because the existing supervised methods and MID-FC(NoPre) have different network
architectures but share similar supervised training schemes with random
initializations, the difference between their performance can illustrate the
contribution of our network.

As shown in \Cref{tab:Net-MID_Valid}, our octree-based HRNet achieves very
similar performance to the best supervised methods in shape classification and
ShapeNet-Part segmentation (with 0.0 and +0.5 differences, respectively). For
fine-grained shape segmentation, the performance of MID-FC(NoPre) is
significantly better (with +11.9 difference). The good performance on different
downstream tasks illustrates that our octree-based HRNet offers an efficient
network architecture for our generic pre-trained model.

\paragraph{Contribution of MID}
For the contribution of the MID loss, we evaluate its contribution in each task
by comparing the performance of MID-FC(NoPre) and MID-FC(Finetune). The two
models share the same network architecture but are trained with different
learning schemes. The MID-FC(NoPre) is trained from scratch with the
task-specific labeled data by a supervised scheme, while the MID-FC(Finetune) is
initialized with the weights of MID-Net pre-trained via our MID scheme and then
refined with the task-specific labeled data.

As shown in \Cref{tab:Net-MID_Valid}, the performance gain of MID-FC(Finetune)
over the supervised MID-FC(NoPre) counterparts in all three tasks (+0.2, +0.6,
+2.4, respectively) clearly demonstrates the contribution of our MID scheme for
different downstream tasks.

\paragraph{Generality of the MID scheme}
To validate the generality and advantage of the MID scheme for pre-training, we
apply the MID scheme to PointNet++ with the default network in \cite{Qi2017}.
Similar to MID-Net, we concatenate PointNet++ with a two-layered FC layer and
fine-tune the networks (denoted as PointNet-FC(Finetune)) for all the tasks. For
comparison, we also train the same concatenated PointNet++ networks from scratch
with the task-specific training data (denote as PointNet-FC(NoPre)).

\Cref{tab:Net-MID_Valid} shows that the performance gain between
PointNet-FC(Finetune) and PointNet-FC(NoPre) in all three tasks exhibits the
efficiency of our MID pre-training scheme to other network architectures, and
the inferior performance of PointNet-FC(Finetune) to MID-FC(Finetune) also
illustrates the importance of network architecture to the pre-trained model and
the advantage of our octree-based HRNet.

\begin{table}[t]
    \centering
\tablestyle{6pt}{1.0}

    \scalebox{0.9}{%
        \begin{tabular}{lccc}
            \toprule
            \multirow{2}{*}{Model} & Shape            & ShapeNet       & PartNet  \\
                                   & Cls.   & Seg.   & Seg. \\
            \midrule
            Prior Art.           & 92.9   & 83.7     & 49.6  \\
            \midrule
            MID-FC(NoPre)        & 92.9                   & 84.2                       & 58.4  \\
            MID-FC(Finetune)     & 93.1                   & 84.8                       & 60.8     \\
            \midrule
            PointNet-FC(NoPre)      &  90.1                  & 81.9                       & 43.8\\
            PointNet-FC(Finetune)   &  90.1                  & 83.4                       & 49.4\\
            \bottomrule
        \end{tabular}
    } 
    \vspace{-5pt}
    \caption{
        Evaluation of our network and the MID scheme. 
    }
    \vspace{-10pt}

    \label{tab:Net-MID_Valid}
\end{table} 

\subsection{Ablation Study}
We pre-train our networks with one modified component and keep all other
settings unchanged.  After pre-training, we test the modified models with shape
classification and ShapeNet-Part segmentation tasks described in the last
section with MID-FC(Fix). To better evaluate the efficacy of our pre-training in
ablation studies, we use a denser version of point clouds provided by
\cite{Wang2017} for the ShapeNet-Part segmentation task which has a similar
point density to the data used in the pre-training. 

\paragraph{HRNet versus other network structures}
Compared with the U-Net, the key advantages of HRNet are the parallel
multi-resolution subnetworks and feature fusion in different resolutions. We
develop two alternative networks that gradually remove the fusion layers between
subnetworks.  The MID-1Fusion removes the first fusion layer from HRNet, while
the MID-NoFusion removes all fusion layers in HRNet.  

As shown in \Cref{tab:ablation}, the accuracy of the network decreases as we drop
more fusion layers (\eg, decrease 0.4\% for MID-1Fusion and 2.1\% for
MID-NoFusion in shape classification), which clearly demonstrates the
importance of the feature fusion. And the accuracy of U-Net drops 1.4\% in
shape classification and 0.5\% in shape segmentation, which validates the
advantage of HRNet in 3D pre-training.

\paragraph{MID Loss versus single-level Loss}
To demonstrate the advantages of MID loss, we train two networks, each of which
is trained with one loss function only. Compared with our original MID loss, the
performance of these two networks (denoted as Shape-Loss and Point-Loss in
\Cref{tab:ablation}) drops, which indicates that each loss function makes its
own contribution to the full network training and affects the performance of
shape encoding in both shape and point levels.

When the training data is limited, the performance gap is further enlarged:
with only 1\% of the training data, the testing accuracy on the classification
task and category IoU on the segmentation task of our network trained with two
loss functions are 61.8 and 74.6, respectively; with only the shape loss, the
accuracy and IoU drop to 60.5 and 61.6; with only the point loss, the accuracy
and IoU drop to 54.4 and 72.2.

\begin{table}
\centering
\tablestyle{6pt}{1.0}
\scalebox{0.9}{
\begin{tabular}{lcc||lcc}
  \toprule
  Model           & Cls.     & Seg.     & Model          & Cls.     & Seg.     \\
  \midrule
  MID-FC          & 90.3     & 85.5     & NoAug          & 82.3     & 83.9     \\
  \cmidrule{1-3} 
  MID-1Fusion     & 89.9     & 85.3     & NoScale        & 89.2     & 84.3     \\
  MID-NoFusion    & 88.2     & 83.1     & NoTrans        & 89.7     & 85.1     \\
  U-Net           & 89.3     & 84.5     & NoRot          & 89.9     & 85.3     \\
  \midrule
  Point-Loss      & 89.5     & 85.3     & 50-Patch       & 90.1     & 84.9     \\
  Shape-Loss      & 88.9     & 83.3     & 200-Patch      & 90.0     & 85.5     \\
                  &          &          & 400-Patch      & 90.2     & 85.4     \\
  \bottomrule
\end{tabular}
} 
\vspace{-5pt}
\caption{Ablation study of different design choices.}
\vspace{-10pt}
\label{tab:ablation}
\end{table}

\paragraph{Augmentation scheme}
We train the network with three different augmentation schemes, each of which
drops one kind of transformation from the original augmentation scheme,
respectively. Compared with the MID-Net trained with full augmentation, the
performances of three networks trained with new augmentation schemes (denoted as
NoRot, NoScale, NoTrans, and NoAug) shown in \Cref{tab:ablation} decrease.

\paragraph{Patch number}
We train the networks with different $K$. As shown in \Cref{tab:ablation}, as
the number of clusters increases from 100 to 400, the performance increases less
than 0.2\% in both shape segmentation and classification tests. We thus set $K$
as 100 in our current implementation to achieve a good balance between training
cost and model performance.

\section{Conclusion}
\label{sec:conclusion}


We propose an unsupervised pre-training method for learning a generic backbone
network from unlabeled 3D shape collections for shape analysis. We
design an octree-based HRNet as backbone network architecture and a
simple-yet-efficient MID loss for pre-training. The ablation study validates the
advantages of joint shape and point feature encoding and training enabled by our
design in the unsupervised pre-training and downstream shape analysis tasks. Our
MID-Net offers state-of-the-art performance for various downstream shape
analysis tasks, especially for tasks with  small training sets.

\bibliography{src/ref/reference}

\appendix

\begin{figure*}
  \centering
  \begin{overpic}[width=0.95\linewidth]{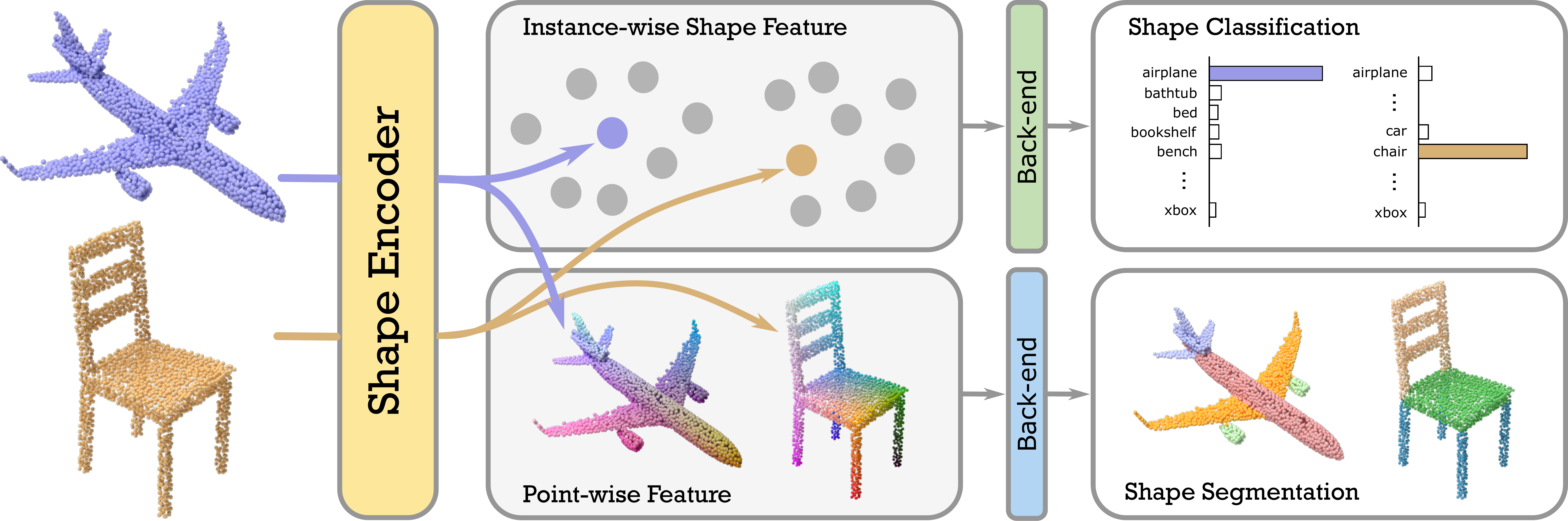}
  \end{overpic}
  \caption{Summary of our multiresolution-instance-discrimination-based learning
  approach. Our MID encoder is learned from a collection of unlabeled 3D shapes.
  It takes 3D point cloud as inputs and generates both instance-wise shape
  features and point-wise features (shown in the middle), which can be used in
  different downstream shape analysis tasks, via concatenated simple back-ends
  with or without fine-tuning.}
  \label{fig:teaser}
\end{figure*}

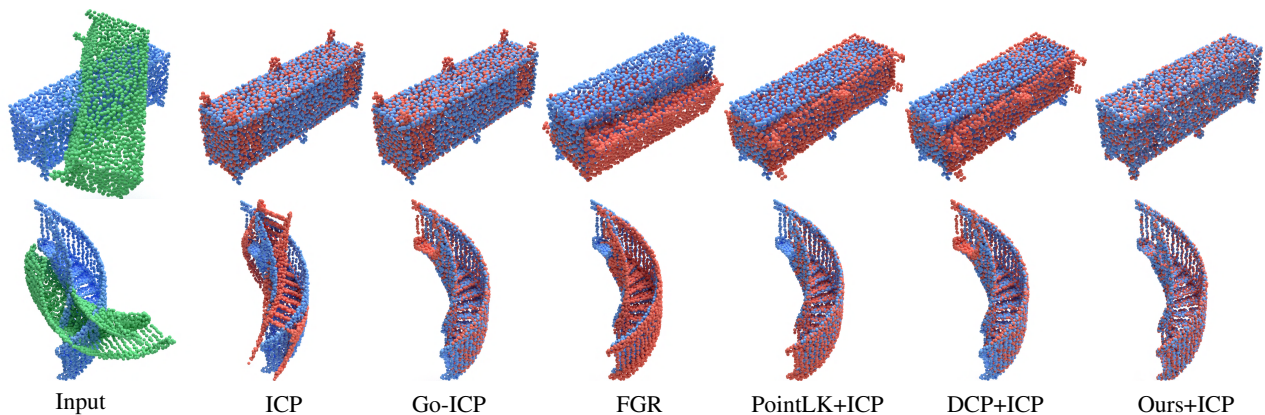
\begin{figure*}
  \centering
  \begin{overpic}[width=0.95\linewidth]{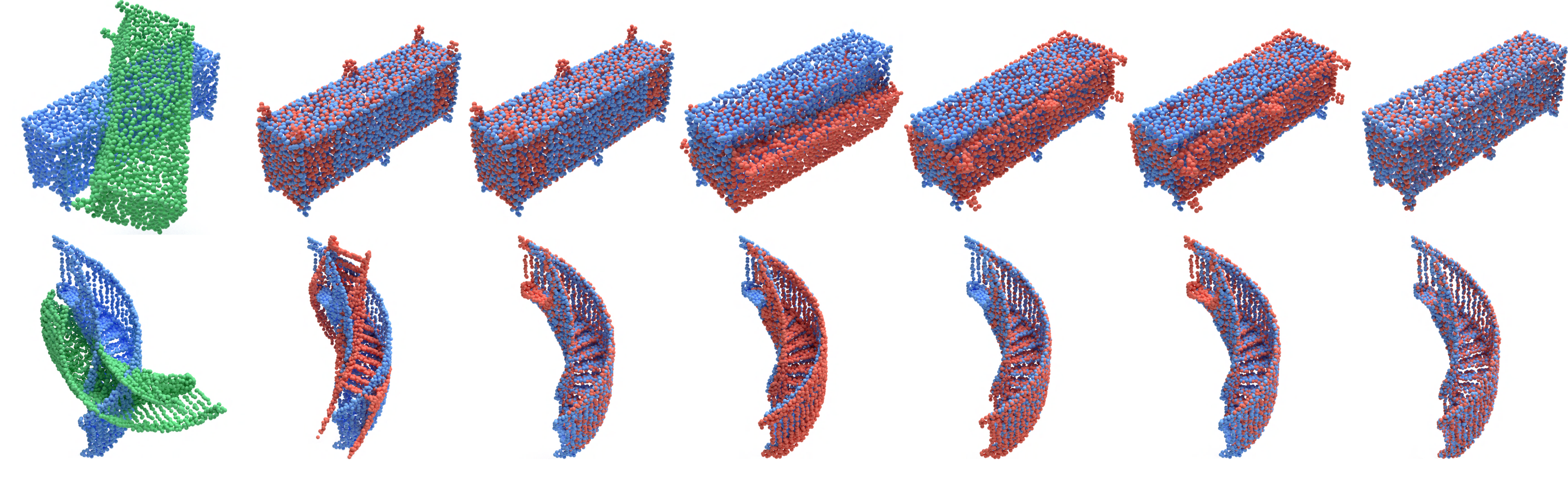}
    \put(7, 0){\makebox(0,0){\small Input}} \put(23, 0){\makebox(0,0){\small
    ICP}} \put(36, 0){\makebox(0,0){\small Go-ICP}} \put(51,
    0){\makebox(0,0){\small FGR}} \put(65, 0){\makebox(0,0){\small
    PointLK+ICP}} \put(79, 0){\makebox(0,0){\small DCP+ICP}} \put(94,
    0){\makebox(0,0){\small Ours+ICP}}
  \end{overpic}
  \caption{Two examples of point cloud registration. The blue shape is fixed,
   the green one is transformed by different methods and its transformed
   version is rendered in orange color.  The stair model is not in the
   ShapeNet dataset. For both examples, only our method registers the inputs
   correctly.}
  \label{fig:regeg}
\end{figure*}

\begin{table*}
  \centering
  \scalebox{0.7}{
  \begin{tabular}{lcccccccccccccccccc}
  \toprule
  Model           & mIoU &Bed  &Bottle & Chair & Clock & Dish & Display & \textbf{Door} & Ear & Faucet & Knife & Lamp & Micro & \textbf{Frid.} & \textbf{Stora.} & Table & Trash     & Vase \\
  \#Train         &      &133  &315  &4489 &406 &111 &633 &149 &147 &435 &221 &1554 &133 &136 &1588 &5707 &221 &741 \\
  \#Test          &      &37   &84   &1217 &98  &51 &191 &51 &53 &132 &77 &419 &39 &31 &451 &1668 &63 &233 \\
  \midrule
  MID-FC(NoPre)   & 58.4 &45.9 &56.1 &54.6 &52.9 &68.1 &90.6 &51.1 &52.1 &64.9 &54.4 &29.9 &75.9 &58.9 &62.8 &43.7 &62.4 &68.3 \\
  MID-FC(Fix)     & 49.4 &42.9 &35.0 &36.2 &44.4 &70.0 &87.5 &48.0 &49.2 &54.1 &46.6 &23.0 &69.0 &50.3 &41.9 &24.7 &55.9 &60.9 \\
  MID-FC(Finetune)& 60.8 &51.6 &56.5 &55.7 &55.3 &75.6 &91.3 &56.6 &53.8 &64.6 &55.4 &31.2 &78.7 &63.1 &62.8 &45.7 &65.8 &69.3 \\
  \bottomrule
  \end{tabular}
  } 
  \vspace{-5pt}
  \caption{Category-specific fine-grained semantic segmentation IoU on
  PartNet. mIoU is the averaged IoU over categories. The categories
  \textbf{Door}, \textbf{Frid.}, and \textbf{Stora.} do not exists in the
  ShapeNetCore55 \cite{Chang2015}. Our network is not trained on  these
  categories even in the unsupervised training stage, and in the finetuning
  stage our network still get improvements.  }
  \label{tab:partnet_full}
\end{table*}

\begin{table*}
  \centering
  \scalebox{0.65}{%
    \begin{tabular}{l|c|cccccccccccccccccc}
      \toprule
      Model & \%train &  C.mIOU & I.mIOU & Aero & Bag & Cap & Car & Chair & Ear. & Guitar & Knife & Lamp & Laptop & Motor & Mug & Pistol & Rocket & Skate. & Table \\
      \midrule     
      MID-FC(NoPre)   &       & 84.1 & 85.2 & 83.4 & 84.7 & 88.1 & 82.0 & 90.3 & 79.1 & 91.1 & 86.9 & 79.6 & 95.4 & 78.0 & 95.9 & 83.3 & 62.9 & 82.7 & 82.5 \\
      MID-FC(Fix)     & 100\% & 82.9 & 84.1 & 80.1 & 83.8 & 90.1 & 78.7 & 89.7 & 80.6 & 90.8 & 85.8 & 79.2 & 95.7 & 71.3 & 93.8 & 80.4 & 62.1 & 81.7 & 82.2 \\
      MID-FC(Finetune)&       & 84.3 & 85.5 & 83.7 & 82.2 & 89.6 & 81.6 & 90.1 & 80.9 & 91.5 & 86.4 & 80.5 & 95.8 & 78.5 & 95.9 & 83.0 & 64.5 & 81.3 & 83.2 \\
      \midrule    
      MID-FC(Fix)     & \multirow{2}{*}{1\%}    & 66.2 & 76.6 & 71.4 & 44.9 & 66.7 & 68.1 & 86.2 & 58.8 & 88.6 & 76.9 & 61.9 & 93.6 & 34.4 & 88.6 & 66.3 & 31.2 & 45.1 & 77.4 \\      
      MID-FC(Finetune)&                         & 67.8 & 76.2 & 73.3 & 60.9 & 80.8 & 71.9 & 86.4 & 53.5 & 89.3 & 72.1 & 57.4 & 93.0 & 21.4 & 80.9 & 67.9 & 42.3 & 56.5 & 77.9 \\
      \midrule
      MID-FC(Fix)     & \multirow{2}{*}{5\%}  & 76.5 & 80.9 & 76.0 & 74.5 & 82.1 & 74.8 & 87.4 & 62.4 & 89.7 & 84.1 & 73.0 & 95.0 & 61.4 & 92.5 & 72.2 & 39.7 & 79.3 & 80.2 \\
      MID-FC(Finetune)&                       & 77.8 & 82.1 & 79.1 & 72.4 & 82.8 & 77.1 & 87.9 & 66.0 & 90.5 & 85.6 & 74.8 & 94.9 & 67.7 & 93.4 & 73.3 & 38.9 & 80.5 & 80.7 \\  
      \bottomrule
    \end{tabular}
  }
  \vspace{-5pt}
  \caption{Category-specific semantic segmentation on the ShapeNet-Part dataset.}
  \label{tab:shapenet_seg_full}
\end{table*}

\section{Training details}
In this section, we provide the training details in our pretraining, and
downstream tasks.

\paragraph{Hyperparameters of the pretraining}
In the pretraining, we set the batch size as 32, momentum as 0.9, and weight
decay as 0.0005. The initial learning rate is 0.03 and it decays by a factor of
10 after 200 epochs and 300 epochs, respectively. The whole training process is
finished after 400 epochs, which takes about 60 hours on an NVIDIA 2080 Ti
graphics card. 

\paragraph{Hyperparameters for the classification task}
For the classification on ModelNet40 dataset \cite{Wu2015}, we optimize the
networks via SGD with 240 epochs. The batch size is 32. The initial learning
rate is set as 0.1 for MID-FC(NoPre) and the FC back-end, and 0.01 for
MID-FC(Finetune). The learning rate  decays by a factor of 1/10 after 120 and
180 epochs.

\paragraph{Hyperparameters for the semantic segmentation task}
In the ShapeNet part segmentation \cite{Yi2016} and PartNet segmentation
\cite{Mo2019}, we used the same set of hyperparameters for training and
finetuning. Specifically, we optimize our networks via SGD with batch size 32.
We train one network for each category seperately. When the number of training
data is large than 2k, we train the network for about 200 epochs; when the
number of training data is less than 1k and 0.5k, we train the network by about
400 epochs and 600 epochs. The initial learning rate is set as 0.1 for
MID-FC(NoPre) and the FC back-end, and 0.01 for MID-FC(Finetune). The learning
rate  decays by a factor of 1/10 after 05\% and 75\% of the total epochs. 

\paragraph{Input signal}
The input to our pre-training network is the point cloud of a shape. We
pre-process the point cloud to assign a normal vector for each point via
principal component analysis if the accurate normal information is not available
in the dataset. For each non-empty octant at the finest level, we fit a plane to
the points inside it, where the plane normal is fixed to the average normal of
these points. We take the unit normal of the plane and the plane offset to the
octant center as the raw feature vector. To deal with the possible inconsistent
normal orientation issue across the shape, we convert each normal component to
its absolute value, and find this trick does not hurt network performance.

\paragraph{Additional details for the point cloud registration}
We uniformly sample 2,048 points from the original shape and the same number of
points from the transformed shapes and take these two point clouds as input for
the shape registration task. We use the ShapeNet as training set, and the
testing set of ModelNet40 as the testing set. As the transformation is known
during the data creation, we can compute the ground-truth registration error by
transforming the two paired shapes together and computing the Hausdorff distance
of these two point clouds.

\section{Visualization}

\begin{figure}[t]
  \centering
  \begin{overpic}[width=1\linewidth]{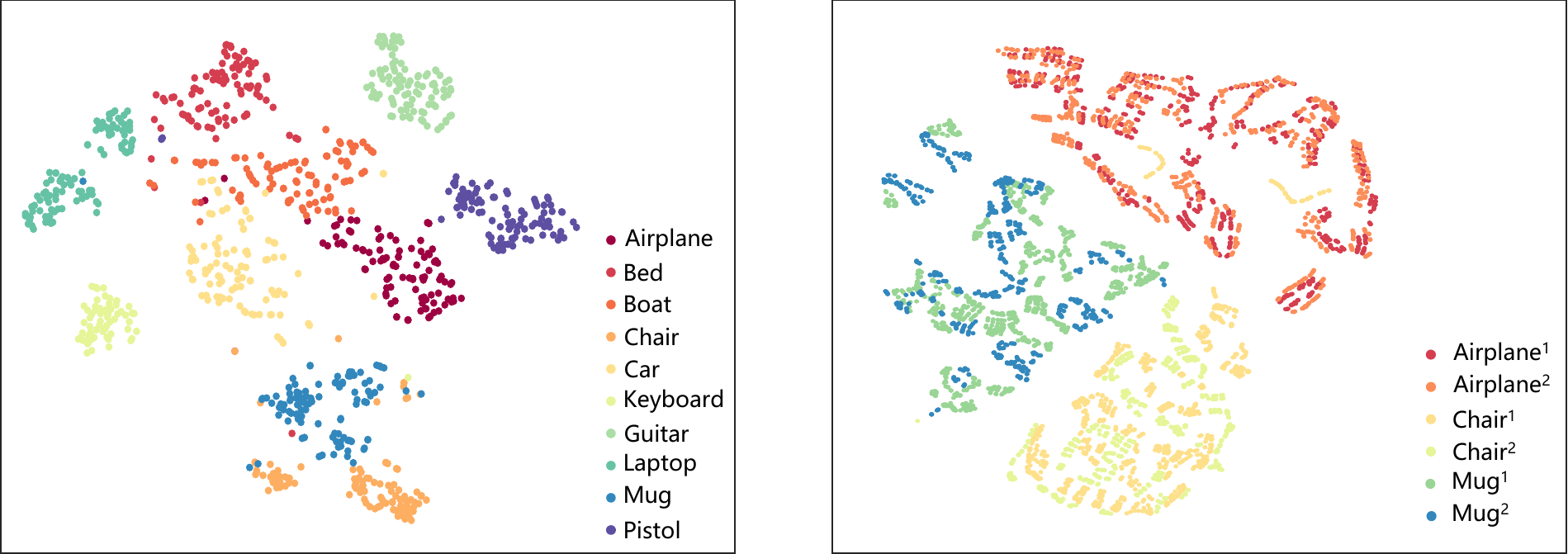}
  \end{overpic}
  \caption{Visualization of t-SNE mapping of shape-level features and point-wise
    features of 3D shapes. The left figure displays the T-SNE map of shape-level
    features 100 3D models selected from 10 shape categories. The right figure
    exhibits the T-SNE map of point-wise features of 6 shapes. }
  \label{fig:tsne}
\end{figure}

\begin{figure}[t]
\centering
\begin{overpic}[width=\linewidth]{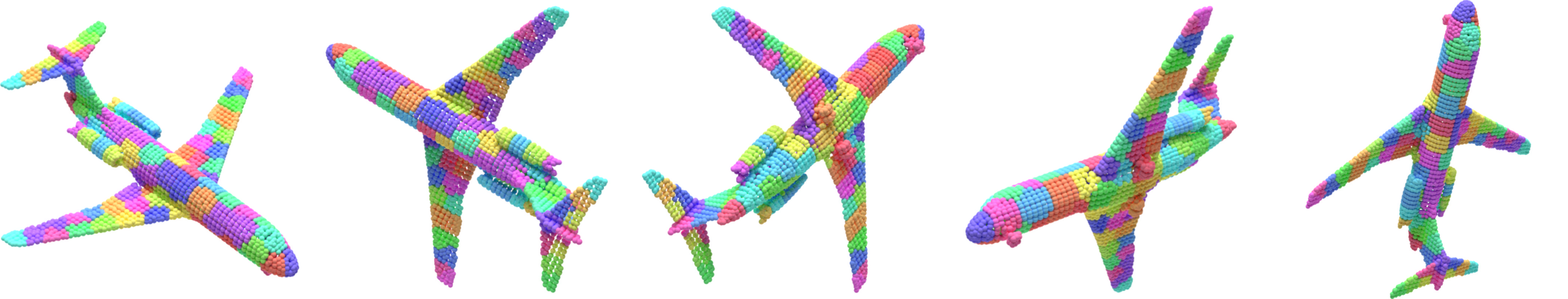}
\end{overpic}
\caption{Visualization of patch-instance classes. An airplane model under
    different transformations is shown here, the over-segmented patches are
    color-coded according to their corresponding IDs. These transformed planes
    also belong to one shape-instance class.} \label{fig:data} 
\end{figure}

\begin{figure}
  \centering
  \begin{overpic}[width=\linewidth]{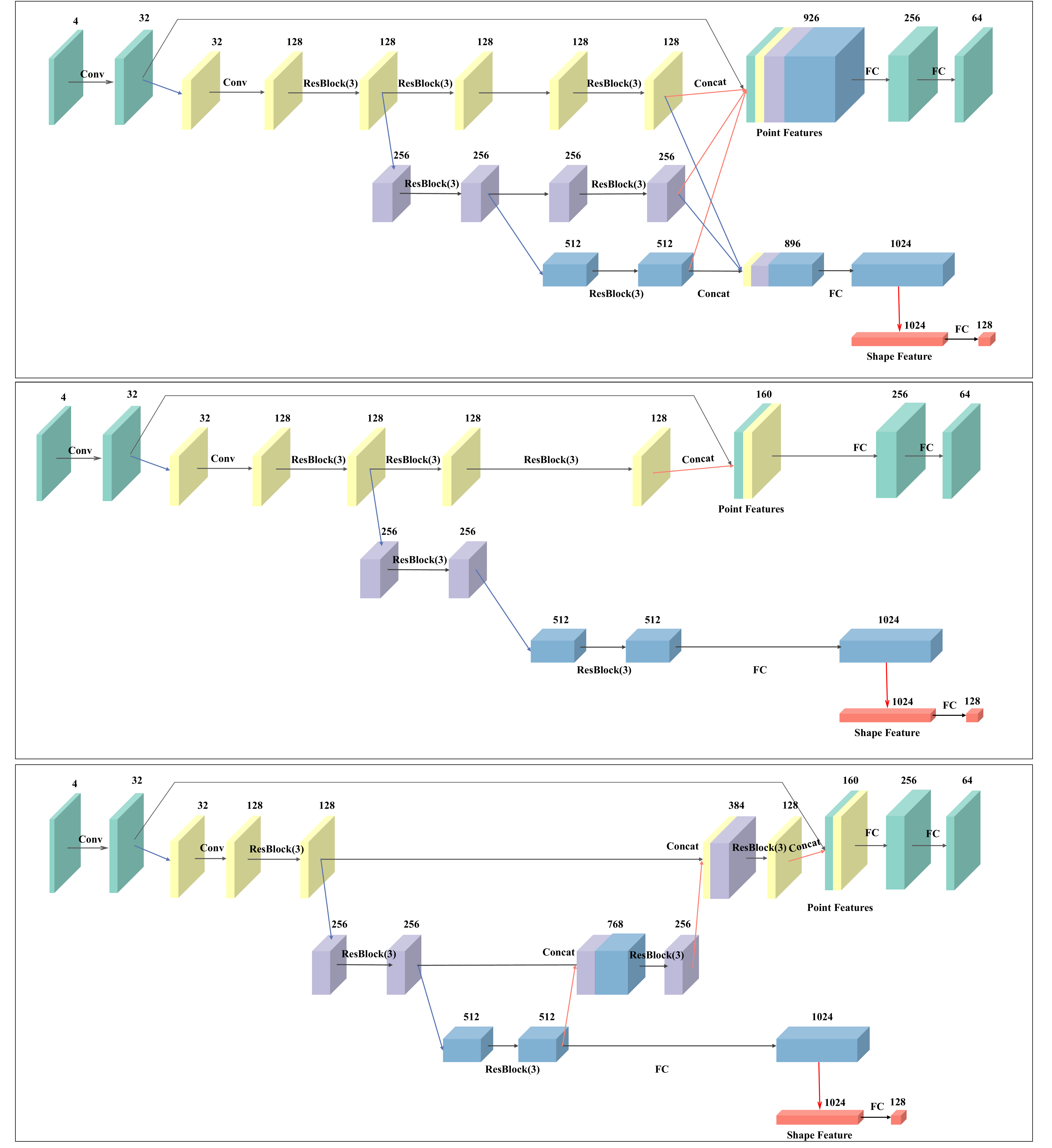}
    \put(5,78){MID-1Fusion} \put(5,43){MID-NoFusion} \put(10,10){U-Net}
  \end{overpic}
  \caption{  Alternative MID network structures for ablation study.   }
    \label{fig:ablation_network}
\end{figure}

\paragraph{Feature visualization}
The quality of learned shape-level features and point-wise features can be
assessed intuitively via visualization. In \Cref{fig:tsne} we show the t-SNE
mapping~\cite{Maaten2008} of shape-level features of 1000 3D shapes randomly
sampled from 10 shape categories, as well as the t-SNE mapping of point-wise
features of 6 shapes (2 airplanes, 2 chairs, and 2 mugs). Note that the
shape-level features and point-wise features of 3D shapes have similar
distributions in their feature space, where the features of 3D models with
similar overall shapes are clustered together and can be easily discriminated
from the features of different 3D models.

\paragraph{Data visualization}
In \Cref{fig:data}, we visualize the segmented patches of an airplane shape and
its augmented copies under different transformations and colorize patches
according to their patch-instance IDs.

\paragraph{Network structure}
In the ablation study of our paper,  we designed two alternative networks
(MID-1Fusion and MID-NoFusion) that gradually remove the fusion layers between
subnetworks. Their network structures, as well as the U-Net, are shown in
\Cref{fig:ablation_network}.

\section{More results}
We provide more detailed results in this section. Specifically, in
\Cref{tab:partnet_full} we show the category-specific segmentation mIoU on the
PartNet dataset; in \Cref{tab:shapenet_seg_full}, we show category-specific semantic
segmentation on the ShapeNet Part dataset; in \Cref{fig:regeg}, we show two
examples of point cloud registration.

\end{document}